\documentclass[10pt,twocolumn,letterpaper]{article}

\usepackage{iccv}
\usepackage{times}
\usepackage{epsfig}
\usepackage{graphicx}
\usepackage{amsfonts}
\usepackage{amsmath}
\usepackage{amsthm}
\usepackage{amssymb}
\usepackage{booktabs}
\usepackage{xcolor}
\usepackage{comment}
\usepackage{color}
\usepackage{epsfig}
\usepackage{pifont}
\usepackage{hhline}
\usepackage{multirow}
\usepackage{booktabs}
\usepackage{algorithm}
\usepackage{algorithmic}

\usepackage[pagebackref=true,breaklinks=true,colorlinks,bookmarks=false]{hyperref}
\usepackage{colortbl}
\definecolor{mygray1}{gray}{.95}
\definecolor{mygray2}{gray}{.9}
\definecolor{mygray3}{gray}{.85}

\iccvfinalcopy 

\def\iccvPaperID{2427} 

\ificcvfinal\fi

\begin{document}

\title{Decoupled Spatial Temporal Graphs for Generic Visual Grounding}



\author{Qianyu Feng$^{1}$,\hspace{2mm} Yunchao Wei$^{1}$,\hspace{2mm} 
Mingming Cheng$^{2}$,\hspace{2mm} 
Yi Yang$^{1}$ \vspace{0.2cm} \\ 
$^1$ReLER, AAII, University of Technology Sydney\\
$^2$Nankai University
}


\maketitle
\ificcvfinal\thispagestyle{empty}\fi
 
\begin{abstract}
 Visual grounding is a long-lasting problem in vision-language understanding due to its diversity and complexity. Current practices concentrate mostly on performing visual grounding in still images or well-trimmed video clips. This work, on the other hand, investigates into a more general setting, generic visual grounding, aiming to mine all the objects satisfying the given expression, which is more challenging yet practical in real-world scenarios. 
 Importantly, grounding results are expected to accurately localize targets in both space and time. Whereas, it is tricky to make trade-offs between the appearance and motion features. In real scenarios, model tends to fail in distinguishing distractors with similar attributes.
 Motivated by these considerations, we propose a simple yet effective approach, named \textbf{DSTG}, which commits to 1) decomposing the spatial and temporal representations to collect all-sided cues for precise grounding; 2) enhancing the discriminativeness from distractors and the temporal consistency with a contrastive learning routing strategy. We further elaborate a new video dataset, GVG, that consists of challenging referring cases with far-ranging videos. Empirical experiments well demonstrate the superiority of DSTG over state-of-the-art on Charades-STA, ActivityNet-Caption and GVG datasets. Code and dataset will be made available.
\end{abstract}


\begin{figure}[t]
    \centering
    \includegraphics[width=\hsize]{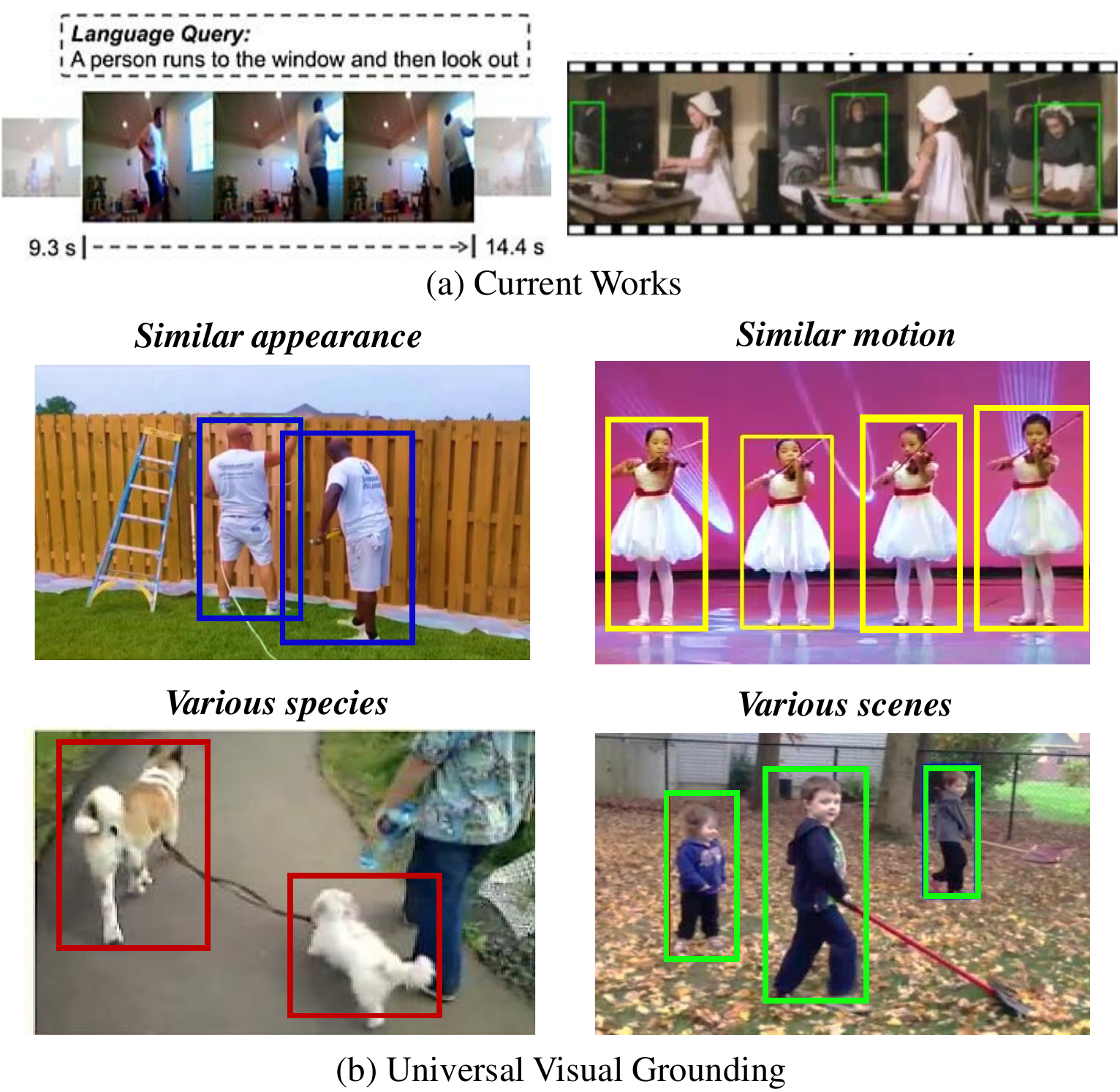}
    \caption{Comparison between current datasets with frames sampled from the generic visual grounding cases. (a) Traditional tasks target a single object in one segment with barely change in the appearance, motion and context; (b) In our GVG, challenging cases with diverse species and scenes are collected, where exist targets possessing similar features in appearance and motion.
    }
    \label{fig:teaser}
\end{figure}

\section{Introduction}
Given a query, \ie, language expression, visual grounding aims at localizing regions or segments via a comprehensive understanding of visual content. It is fundamental to many potential applications such as human-computer interaction~\cite{conf/cvpr/LiaoLWCQF20,tang2020humancentric}, robotics and video surveillance~\cite{Huang_2019_ICCV,dalal2006human}. 
Due to its huge prospects, visual grounding has drawn much attention~\cite{gao2017tall,gavrilyuk2018actor,krishna2017dense,wiriyathammabhum2019referring} recently.
At first, targets are queried in the images based on the category, attributes and context. 
In real-life scenarios, performing visual grounding in both space and time is a more practical, reasonable yet challenging task. 
Compared to image-based grounding, video-based grounding leaps the gap off the page to real life by taking account of temporal information in the frame sequence. Previous works~\cite{luo2017comprehension,nagaraja2016modeling} focus on jointly learning the encoded embeddings of the object and language. 
However, current efforts~\cite{regneri2013grounding,gao2017tall,hendricks17iccv} only consider the localization in the space or at temporal level, \ie, the segment of an action or described event, rather than pointing out the spatial regions related. 
Most recently, a rising number of works \cite{gu2018ava,shang2019annotating,zhou2019grounded} are dedicated to tackling visual grounding by considering both spatial and temporal information. The key to solving this task is to capture the critical features of objects that corresponded to referring expressions.

It is noteworthy that existing works~\cite{kazemzadeh2014referitgame,regneri2013grounding,gao2017tall} fail to meet the general requirements of visual grounding in real scenarios. There is little room left for models to develop in most referring cases, \eg, only one or two objects exist in video while the target dominates the frame. Besides, with a limited scale and diversity, actions and activities tend to be attached to a certain context. With the bias in the annotation, models tend to focus on the salient attributes while neglecting the critical parts and thus fail to generalize into real application. It is usually hard to distinguish targets from appearance-affinitive or motion-affinitive distractors, especially in video with group activities, \eg, \textit{dancing, playing the football, skiing}. Moreover, the entanglement of motion with the scene makes it harder to ground targets, \eg, visual features of \textit{a man with swimsuit near a pool} will confuse the model to attach it to \textit{swimming}. 

So, what characteristics should a generic visual grounding task possess? \textbf{On one side}, it should be placed under the untrimmed setting. The real-world is naturally continuous. Given a referring expression, we should localize the targets within a long untrimmed video by excluding the unrelated chaotic content in space and time. Rather than simply retrieving detected regions in well-trimmed videos, algorithms should be equipped with potent abilities for comprehension and localization. 
\textbf{On the other side}, referring cases should be as diverse as possible to approach real-life scenarios. For instance, one referring expression might correspond to multiple segments or multiple objects rather than the singular pair-wise referring cases composing the current related datasets.  

To remedy the above issues, we propose a new method, Decoupled Spatial Temporal Graphs (DSTG), to learn a discriminative representation by decomposing the spatial and temporal information. In this way, the model captures not only the spatial appearance but also temporal motion features simultaneously. Targets are more distinguishable from distractors with similar spatial or temporal features.
Additionally, to alleviate the ambiguity from these distractors, a Spatio-Temporal Contrastive Routing (STCR) Strategy is carried out by pairing targets with positive samples while randomly selecting distractors as negative samples to maintain the consistency across space and time. 
To advance the research of visual grounding to a general prototype, we also contribute to a new video dataset, GVG, which incorporates diverse cases and takes a step to approach generic visual grounding. 
To sum up, the contributions of our work are as follows:
\begin{itemize}
    \item We propose a new architecture, DSTG, for visual grounding task by decoupling the spatial and temporal representation with two branches. One focuses on learning the appearance features and relationship with the context. The other captures the attributes for motion and temporal variation.  
    
    \item We put forward a practical setting, generic visual grounding. A STCR Strategy is devised to identify targets from distractors and withstand the temporal variance with robust and expressive representations. Besides, our method is also proposal-free temporally which is more flexible and tractable.
    
    \item Furthermore, we establish a new video benchmark GVG with large diversity. The superiority of the proposed method is demonstrated over state-of-the-art with extensive experiments on Charades-STA, ActivityNet-Caption and GVG datasets.
    
\end{itemize}

\begin{figure*}[t]
    \centering
    \includegraphics[width=0.9\hsize]{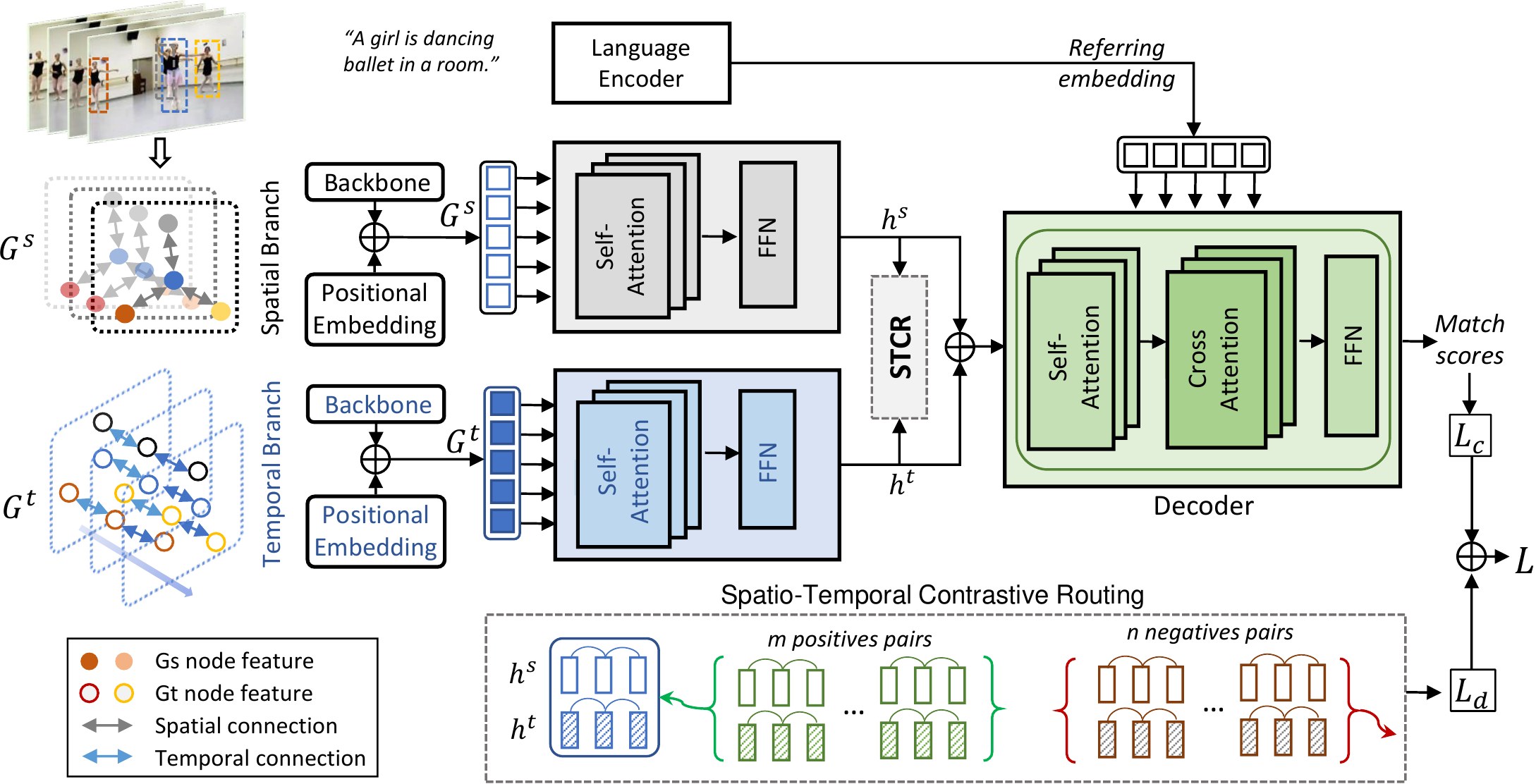}
    \caption{An overview of the proposed method DSTG, which targets on visual grounding by first learning spatial and temporal representations separately to better capture the appearance and motion features. Then, a spatio-temporal contrastive routing strategy is applied to enhance the discriminativeness from distractors and maintain the consistency across time. }
    \label{fig:framework}
    \vspace{-3mm}
\end{figure*}

\section{Related Work}
\noindent \textbf{Visual Grounding.}
Visual grounding task~\cite{plummer2015flickr30k,kazemzadeh2014referitgame,mao2016generation} is first put forward to refer objects in the image with expression in natural language. It is fundamental to other vision-language tasks, \eg, visual question answering~\cite{antol2015vqa,yang2016stacked}, visual reasoning~\cite{clevert2015fast,conf/cvpr/LiaoLWCQF20}, relationship modeling~\cite{lu2016visual,journals/corr/abs-2101-08165}.
The conventional approaches are built on the CNN-RNN structure. 
Furthermore, with the popularization of the attention mechanism, it is also introduced into the grounding tasks~\cite{allport1989visual,xu2015show,vaswani2017attention,endo2017attention,deng2018visual}. Compared to the global representation, local features are more robust to variation in the lighting changes, camera shaking, and noises from the background. 
For a better comprehension of language, some approaches reach out for an external language parser~\cite{conf/cvpr/LiaoLLWCQL20,liu2019learning,journals/corr/abs-2010-00515,conf/eccv/HuiLHLYZH20} which helps to better understand the semantic meanings by breaking down the complex grammar. 
These methods mainly focus on extracting the feature of the objects, or learning with simple hand-crafted relationships between them. For better exploiting the potential relationship as well as the correspondence with the referring expressions, tree-based~\cite{conf/aaai/WuLLL20,journals/corr/abs-2001-06680} and graph-based methods~\cite{wang2019neighbourhood,yang2019dynamic,liu2020learning} take full advantage of learning the embedding of the objects and its neighbors.
 
\noindent \textbf{Temporal Localization.}
The temporal localization tasks~\cite{regneri2013grounding,gao2017tall,anne2017localizing} mainly solve recognition and localization of actions.
Difficulties emerge as the video brings conspicuous ambiguity in the temporal variation with unclear boundaries and redundant representation. 
Approaches~\cite{song2016unsupervised,xu2015jointly,li2020csrl,li2020hierarchical} have been raised to learn a common embedding space from different modalities.
Later, attention~\cite{liu2018attentive,yuan2019find} and modular-based~\cite{zhang2019man} methods are proposed and applied into the temporal localization tasks.
Ge \etal~\cite{ge2019mac} proposed to mine the semantic concepts to enhance the temporal grounding performance. 
Thereafter, Wang \etal~\cite{wang2020temporally} learned to predict the temporal boundary with offsets based on the window anchors.
However, these methods lack the fine-grained understanding of the visual content, thus suffering from the appearance change and noise from the background.

\noindent \textbf{Video Understanding.}
With the rapid development of the multi-modal research, tasks like video captioning~\cite{ma2017grounded,zhou2019grounded}, video navigation~\cite{zhu2017icra} come upon the stage, which requires a comprehensive understanding of both the vision and language.
Recently, numerous works~\cite{tu2017video,Huang_2019_ICCV,shang2019annotating,li2020hmseg,li2020metaparsing,li2021self,li2021superresolving} start to emphasize the significance of the fine-grained understanding of spatial regions in video-related tasks.
AVA~\cite{gu2018ava}, a video benchmark has been proposed to localize actions spatially and temporally.
Most methods~\cite{chen2018tgn,wang2020temporally} are limited with the bottleneck from figuring out the temporal boundaries and indistinguishable from the spatial or temporal distractors.
Casting off the temporal proposals, our work is the first to explore the fine-grained understanding for visual elements.

\section{Method}
Our DSTG tackles the generic visual grounding task for localizing objects in untrimmed videos. We first give a brief introduction to the problem. Then, we elaborate the detailed design in capturing spatial and temporal representations separately. On top of that, we further enhance the consistency by Spatio-Temporal Contrastive Routing, which eliminates the ambiguity from distractors.

\subsection{Problem Statement}
Given a referring expression $\boldsymbol{r}$, visual grounding aims to localize the target from language description. We focus on grounding in videos which is more general.
Taken the input video as a sequence of frames, the goal is to output spatial location and temporal segment of the described object. Specifically, region proposals can be obtained with detection models. 
For each region in a frame, RoI Align~\cite{ren2015faster} can be applied to extract its feature. Here we denote the $N$ region features from a video as $\{{\boldsymbol{x}_i}\}_{i=1}^{N}$. 
The region feature $\boldsymbol{x}$ is the output vector before the final layer of the pretrained model. Besides, we also take the advantage of positional embedding to capture geometric information by concatenating the location coordinates and area size {\small$(x_0, y_0, x_1, y_1, w\cdot h)$} against the frame size {\small$(W, H)$} into the extracted feature, denoted as $\boldsymbol{x} =[\boldsymbol{x}, f(\frac{z_0}{W}, \frac{y_0}{H}, \frac{x_1}{W}, \frac{y_1}{H}, \frac{w \cdot h}{W \cdot H})]$. 

In this work, we investigate into a general setting, generic visual grounding, where the referring expression describes one or more objects in untrimmed videos. Different from the traditional cases with one target in a single period, we focus on more general cases: (1) one object in several discontinuous clips can be referred to by one expression; (2) one object performs different actions in two or more discontinuous clips, and be referred to by two or more expressions accordingly; (3) multiple objects with similar appearance perform the same action, \eg, \textit{dance, sport activities}, thus can be referred to with the same expression. 

\subsection{Decoupled Spatial Temporal Graphs}
We propose a novel framework, DSTG, to tackle the diverse referring cases. 
For grounding target in videos, it is crucial to fully exploit visual cues from the spatial context, \eg, appearance features and relationship with other objects. 
Besides, the motion pattern and dynamic variance matter in precise localization temporally. 
Our core idea is to decompose the representations into spatial and temporal manifolds. 
The overview of our DSTG is shown in Figure ~\ref{fig:framework}, which is dedicated to learning the correspondences of representation in three dimensions: spatial-spatial consistency, temporal-temporal consistency, spatial-temporal-language matching. 

\textbf{Encoder}
At first, we apply an off-the-shelf parser~\cite{bird2009natural} on the expression $\boldsymbol{r}$.
The embedding of each sentence is learned with a recurrent network. 
We propose to learn spatial and temporal representation with a graph-based encoder-decoder architecture.
The spatial graph $\boldsymbol{G}^s$ and the temporal graph $\boldsymbol{G}^t$ are built separately over the region proposals $\boldsymbol{X} = \{\boldsymbol{x}_i\}_{i=1}^{N}$ from the frames.

In the spatial branch, the relationship between each node with its $K_s$ spatial neighbors can be calculated with $e_{ij} = a(\boldsymbol{x}_i, \boldsymbol{x}_j), j\in K_s$, where $a(\cdot, \cdot)$ denotes a linear transformation on the concatenation of $\boldsymbol{x}_i$ and $\boldsymbol{x}_j$. Then, $e_{ij}$ is normalized to calculate the attention weights with:
\begin{equation}
    \alpha_{ij} = \frac{exp(\textit{LeakyReLU}(e_{ij}))}{\sum_{k=1}^{K_s} exp(\textit{LeakyReLU}(e_{ik}))}.
    \label{eq:node_att}
\end{equation}
With the messages aggregated from the neighbors, the embedding of the current node can be updated as:
\begin{equation}
    \boldsymbol{h}_i^s = \sigma({\sum_{j=1}^{K_s} \alpha_{ij} \boldsymbol{x}_j}),
    \label{eq:node_update}
\end{equation}
where $\sigma$ denotes the nonlinear transformation with sigmoid function. 

Parallelly, a temporal graph $\boldsymbol{G}^t$ is built to capture the dynamic characteristics of motion, \eg, pose changes, body moves.
To obtain the motion-related features, we adopt a 3D backbone model~\cite{carreira2017quo} and apply RoI Align on the extracted feature. 
We explore the temporal embedding of each region by connecting it to its $K_t$ temporal neighbors. 
Again, the relationship with its temporal neighbors are calculated with Eq.~(\ref{eq:node_att}). Node features $\boldsymbol{h}_i^t$ in graph $\boldsymbol{G}^t$ can be updated with Eq.~(\ref{eq:node_update}). Finally, each node embedding is the concatenation of spatial and temporal features: $\boldsymbol{h}_i = [\boldsymbol{h}_i^s, \boldsymbol{h}_i^t]$.

\textbf{Decoder}
With the decoupled spatial and temporal representations from the encoder, decoder consists of graph-based self-attention and cross-attention for learning the correspondence between the visual elements with the referring expression.
Intuitively, the encoded graph embeddings serve as the latent representation for candidate instances to match. 
Together with the $N$ nodes referring to region proposals in a video, the language embeddings of referring query $\mathbf{r}$ from language encoder are fed in the decoder. 
We assume that the number of nodes $N$ is fixed and is sufficient for learning the panoramic cues from the spatial and temporal context. 

To learn the interactions among the graph-structured query embeddings, we first apply graph self-attention modules (same as the encoder). 
Additionally, the cross-modal attention module learns interactions between the referring expression and the context graph, \ie, the latent representation of objects, and referring query embeddings, \ie, the latent representations of the language expressions.
We represent the interactions between the node embedding $\boldsymbol{h_i}$ and $\boldsymbol{r}$ by attending the latter over the nodes in the graph. The attention distribution over the nodes with $r$ is calculated as:
\begin{equation}
    \gamma_i = \sigma(\frac{\text{exp}(a(\boldsymbol{h}_i, \boldsymbol{r})}{\sum_{j=1}^N\text{exp}(a(\boldsymbol{h}_j, \boldsymbol{r}))}).
\end{equation}
Overall, the cross-modal attention module models the interactions between the latent representations of the input video and the action queries. The spatial correspondence score can be calculated for region $\boldsymbol{x}_i$ with the attended feature $\hat{\boldsymbol{h}}_i=\gamma_i \boldsymbol{h}_i$ by: 
\begin{equation}
    c(\boldsymbol{x}_i, \boldsymbol{r}) = \sigma(W_h  \hat{\boldsymbol{h}}_i \cdot W_r \boldsymbol{r}),
\end{equation}
where $W_h$ and $W_r$ are parameters to be learned.
With corresponding scores calculated, we can calculate the matching loss between the referring expression $r$ and region $x_i$:
\begin{equation}
    \mathcal{L}_c (\boldsymbol{x}_i, \boldsymbol{r}) = - y\log c(\boldsymbol{x}_i, \boldsymbol{r}) - (1-y)\log (1-c(\boldsymbol{x}_i, \boldsymbol{r})).
\end{equation}
where $y=1$ when $\boldsymbol{x}_i$ is matched with $\boldsymbol{r}$, otherwise, $y=0$.

\subsection{Spatio-Temporal Contrastive Routing}
Current grounding methods focus on the matching of encoded embeddings with the referring expression over all the samples. However, the corner cases are hard to figure out with appearance-affinitive and motion-affinitive distractors. Hence, we elaborate a Spatio-Temporal Contrastive Routing (STCR) strategy by selecting spatial distractors and temporal distractors as negative samples.

It is discovered that objects corresponded with the referring expression are much fewer compared to the negative ones. 
A contrastive loss~\cite{hadsell2006dimensionality, he2020momentum} is proposed to deal with such cases forcing targets to be similar to its positive samples and dissimilar to other negative samples. 
Furthermore, it is non-trivial for the model to keep the consistency in both space and time. 
Hence, we develop a new contrastive learning based mechanism to enhance the distinctiveness of spatial and temporal representation. 
The loss function for measuring the consistency of representation is defined as:
\begin{equation}
\begin{split}
    \mathcal{L}_{d}(\boldsymbol{x}_i) &= \sum_{j=1}^{m} d(\boldsymbol{h}_i^s, \boldsymbol{h}_j^s)  / m + \sum_{k=1}^{n} (1 - d(\boldsymbol{h}_i^s, \boldsymbol{h}_k^s)) / n\\
    &+ \sum_{j=1}^{m} d(\boldsymbol{h}_i^t, \boldsymbol{h}_j^t)  / m + \sum_{k=1}^{n} (1 - d(\boldsymbol{h}_i^t, \boldsymbol{h}_k^t)) / n,
\end{split}
\vspace{-2mm}
\end{equation}
where $m,n$ denote the number of positive and negative samples respectively.
Empirically, the euclidean distance is better than the cosine similarity in measuring the visual distance between two objects~\cite{li2003dist}. In total, our objective function is defined as:
\begin{equation}\vspace{-2mm}
    \mathcal{L}(\boldsymbol{X}, \boldsymbol{r}) = \frac{1}{N}(\sum_{i=1}^{N}\mathcal{L}_c(\boldsymbol{x}, \boldsymbol{r}) + \lambda \mathcal{L}_{d}(\boldsymbol{x}_i)),
    \vspace{-1mm}
\end{equation}
where $\lambda$ is a hyper-parameter to balance the weight of consistency loss.

\begin{figure}[t]
    \centering
    \includegraphics[width=0.95\hsize]{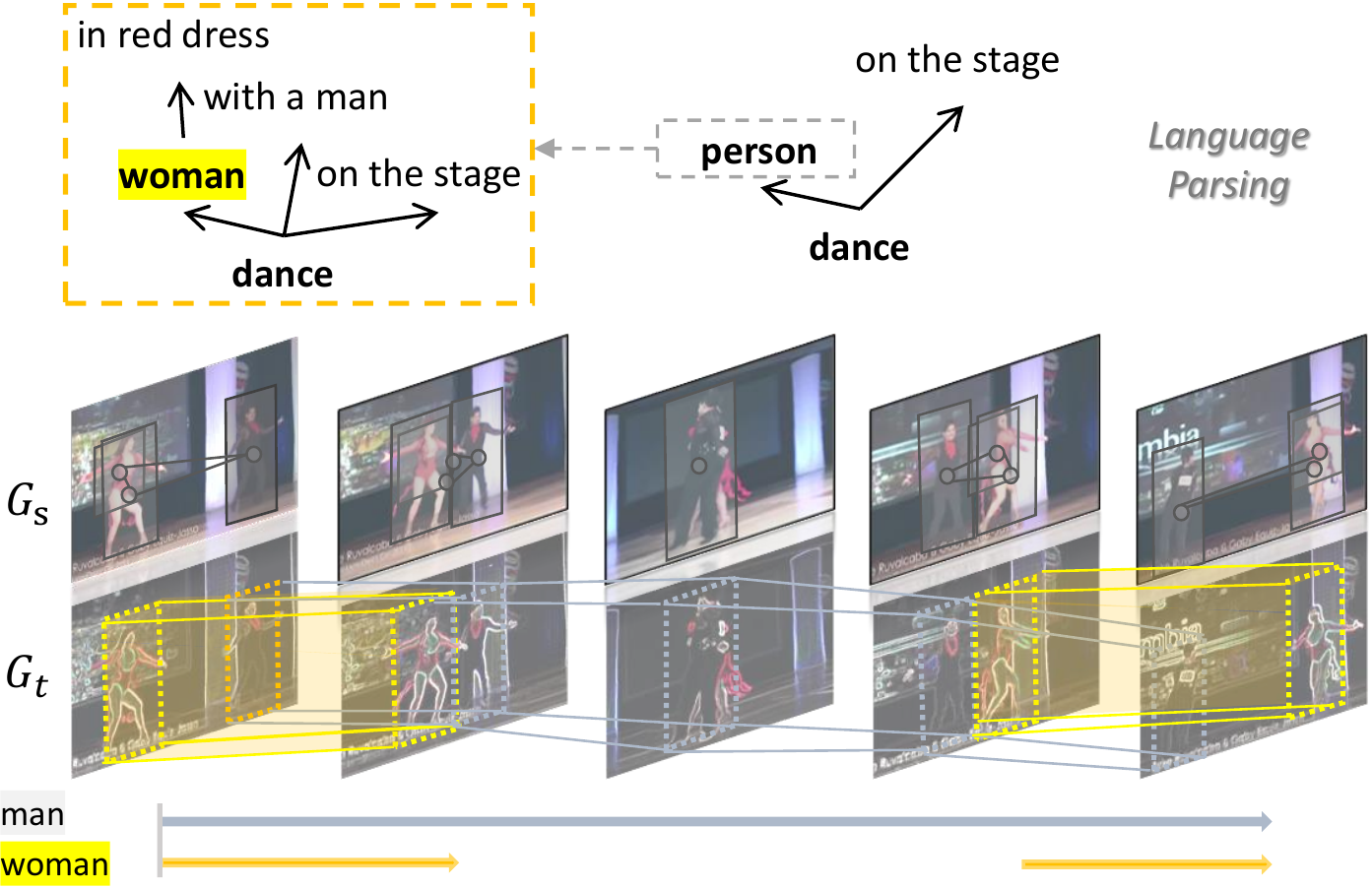}
    \caption{Illustration of the grounding of referring a single object and multiple objects in video.}
    \label{fig:preview}
    \vspace{-3mm}
\end{figure}

By decomposing the spatial and temporal representation, our model learns more distinguishable and robust features with STCR. Therefore, it can better handle challenging referring cases with large diversity and variety. For example, in Figure~\ref{fig:preview}, the spatial feature of ``woman'' is different from the ``man''. So, our method is able to ground the correct target in discontinuous segments. From another perspective, they possess similar motion feature of ``dance''. Both of them will be grounded given expression ``\textit{a person is dancing on the stage}.'' In this manner, targets can be better distinguished from the spatial/temporal distractors with similar attributes. 

\textbf{Inference}
Different from the propose-and-rank approaches adopted by previous work, our method functions in an end-to-end and proposal-free style, which gets rid of stereotyped anchors.
During the inference, we first detect potential region proposals and extract the region features. Then, spatial graph $\boldsymbol{G}^s$ and temporal graph $\boldsymbol{G}^t$ are built and forwarded into the encoder.
The encoded spatial and temporal features are concatenated together and fed into the decoder with the language embeddings for the cross-modal attention to calculate the matching scores.
For each region $\boldsymbol{x}_i$, the reward score for it to be linked to another region $\boldsymbol{x}_j$ (considering two regions belong to the same target) is calculated as:
\begin{equation}
    \begin{aligned}
    R_{ij} =c(\boldsymbol{x}_i, \boldsymbol{r}) + c(\boldsymbol{x}_j, \boldsymbol{r})
     - d(\boldsymbol{h}_i^s, \boldsymbol{h}_j^s) - d(\boldsymbol{h}_i^t, \boldsymbol{h}_j^t) 
    \end{aligned}
    \label{eq:link}
\end{equation}
At last, the tube-level NMS is applied to linked regions for filtering duplicated tubes to generate final grounding results. 

\section{Experiments}
\subsection{Implementation Details}
\noindent\textbf{Training.} For the fine-grained localization, a pretrained Faster R-CNN~\cite{ren2015faster} with ResNet-101 as backbone is adopted as the detection model. 
During the training, the ground truth regions are also adopted.
Then, we apply RoI Align~\cite{ren2015faster} on the extracted features to get the regional features. 
For the referring expression, each word in the sentence is represented by the pretrained word embeddings with GloVe~\cite{pennington2014glove}.
Dropout~\cite{elliott1974delinquency} is applied in the language encoder, bi-directional LSTM, with a ratio of 0.1.
In all the encoder and decoder modules, the graph self-attention layer number is 2 with the residual connection. 
\textit{LeakyReLU} is applied after each graph layer.  
For videos with numerous objects from the same category, we limit the maximum number from the same class to enlarge the diversity of the regions.
In the final objective function, hyper-parameter $\lambda$ is set to 0.2 empirically. 
During the inference, the tube NMS is chosen to be 0.4 empirically.

\noindent\textbf{Evaluation metrics.} 
We systematically evaluate and verify the effectiveness of the proposed method against the state-of-the-art methods. 
For temporal localization on Charades, we applied mAP following the same setting in ~\cite{sigurdsson2016hollywood}. 
On Charades-STA and ActivityNet-Captions, m\_IoU and IoU@0.5 are applied following~\cite{wang2020temporally}.
For fine-grained visual grounding, we employ the m\_vIoU and vIoU@R as the main criteria metrics. Taking $F_U$ as the set of frames contained in the selected or ground truth clips, and $F_I$ as the intersection of the selected and ground truth clips, we calculate vIoU by calculating the sum of the IoU of the predicted regions and ground truth in frame $\in F_I$ averaged over $F_U$. The m\_vIoU is the average vIoU of samples and vIoU$_{@R}$ is the proportion of samples whose vIoU$_{\geqslant R}$. We also compare m\_tIoU, the average temporal IoU for different localization methods.

\begin{table}[t]
    \centering
    \caption{Statistic numbers of different word types counted from the vocabulary of referring expressions in GVG Dataset.}
    \setlength{\tabcolsep}{7.8pt}
    \begin{tabular}{|l|cc|c|} 
    \hline
    Count & Total & Unique & Top-3 \\
    \hline
    \textit{noun}      & 27,463  & 159 & man, person, woman \\
    \textit{verb}      & 18,327  & 342 & play, talk, walk\\
    \textit{adjective} & 25,306  & 562 & red, tall, slow\\
    \hline
    \end{tabular}
    \label{tab:gvg_vocab}
    \vspace{-3mm}
\end{table}

\subsection{GVG Benchmark}
\noindent\textbf{Data collection.}
Considering the limited cases in the existing datasets, we collect a new benchmark with broader diversity in real-life scenarios. To this end, we carefully select videos with adequate length and varieties under both indoor and outdoor scenes with rich species and activities. Then, we annotate the objects involving in the main activities with referring expressions, containing at least one phrase for the appearance or relationship, and one phrase for the action or activity. 

\noindent\textbf{Data statistics.}
In total, the GVG benchmark consists of 5,537 videos and 16,512 referring expression sentences. The lengths of the collected videos vary from 16s to 223s. The average length of the videos is 124.5s. The referring sentences consist 5 to 22 words with an average length of 9.8 words. 
There are 3,814 out of 5,537 videos with multiple referents. The category of object is 46 and the most is `man' with 2,316 cases and the least is `camel' with 12 cases in the 16,512 referring samples.
We split the GVG dataset as 11,625 / 2,474 / 2,413 paired samples as train, validation and test set respectively. 
More details can be found in the Appendix.

\subsection{Empirical Study}

\subsubsection{Temporal Localization}
\noindent \textbf{Charades-STA} is a video dataset containing 9848 clips of daily indoor activities. 
Charades-STA is built on Charades~\cite{sigurdsson2016hollywood} dataset, which focuses on indoor activities with 157 action categories. Videos in the dataset contain an average of 6 action instances per video with an average of 79\% of overlapping instances in a video. This dataset is challenging because of the high degree of overlap in the action instances. The temporal annotations of Charades-STA are generated in a semi-automatic way, which involve sentence decomposition, keyword matching, and human check. The videos are 30 seconds on average. The train/test split is 12,408/3,720. 

\noindent \textbf{ActivityNet Captions} is built on \emph{ActivityNet} v1.3 dataset with averaged 2-min videos. This dataset has much larger variation, ranging from several seconds to over 3 minutes. We take the two validation subsets as the test split following~\cite{chen2018tgn}. The numbers of query-segment pairs for train/test split are thus 37,421 and 34,536.
Following~\cite{chen2018tgn}, we respectively set the anchor number to 20 and 100 for Charades-STA and ActivityNet Captions. 

\noindent\textbf{Performance.}
We evaluate the temporal localization performances with the state-of-the-art methods in Table~\ref{tab:activity_sota} to show the advantage of our DSTG. For a fair comparison, we apply C3D network~\cite{tran2015learning} to extract the regional features in videos. 
On Charades-STA, our method reaches 36.7 for mIoU. DSTG outperforms CBP by 1.0 for mIoU on Charades-STA and 0.4 on ActivityNet-Cap. 
It is discovered that with STCR module, DSTG learns more discriminative features with performance increasing mIoU from 35.2 to 36.7. Whereas, the anchor-based methods are ultimately restrictive in scope since they are scaffolded with the predefined clips as candidates, which makes it hard to extend for videos with considerable variance in length.





\setlength{\tabcolsep}{7.7pt}
\renewcommand{\arraystretch}{0.95}
\begin{table}[t]
\centering
\caption{Comparison with state-of-the-art methods on Charades-STA and ActivityNet-Captions with IoU$_{@0.5}$ and mIoU scores reported.}
\scalebox{0.9}{
\begin{tabular}{l cccc}
\toprule
\multirow{2}{*}{method} & \multicolumn{2}{c}{\begin{tabular}[c]{@{}c@{}} Charades-STA\end{tabular}} &
    \multicolumn{2}{c}{\begin{tabular}[c]{@{}c@{}} ActivityNet-Cap\end{tabular}}   \\ \cline{2-5}
 & IoU$_{@0.5}$ & mIoU & IoU$_{@0.5}$ & mIoU\\

\midrule
Random Anchor & 14.6 & 20.4 & 13.3 & 18.4\\
CTRL ~\cite{gao2017tall} & 21.4 & - & - & - \\
TGN ~\cite{chen2018tgn} & - & - & 27.9 & 29.2\\
TripNet ~\cite{piergiovanni2018learning} & 22.3 & - & 32.2 & -\\
Xu \etal ~\cite{mavroudi2020representation} & 36.6 & -& 27.7 & -\\
CBP~\cite{tang2013combining} & 36.8 & 35.7 & 35.8 & 36.9\\

\midrule
DSTG w/o STCR & 36.5 & 35.2 & 36.1 & 36.7 \\
\textbf{DSTG (Ours)} & \textbf{37.2} & \textbf{36.7} & \textbf{36.6}& \textbf{37.3}\\

\bottomrule
\end{tabular}}
\label{tab:activity_sota}
\end{table}

\begin{figure}[t]
    \centering
    \includegraphics[width=1.02\hsize]{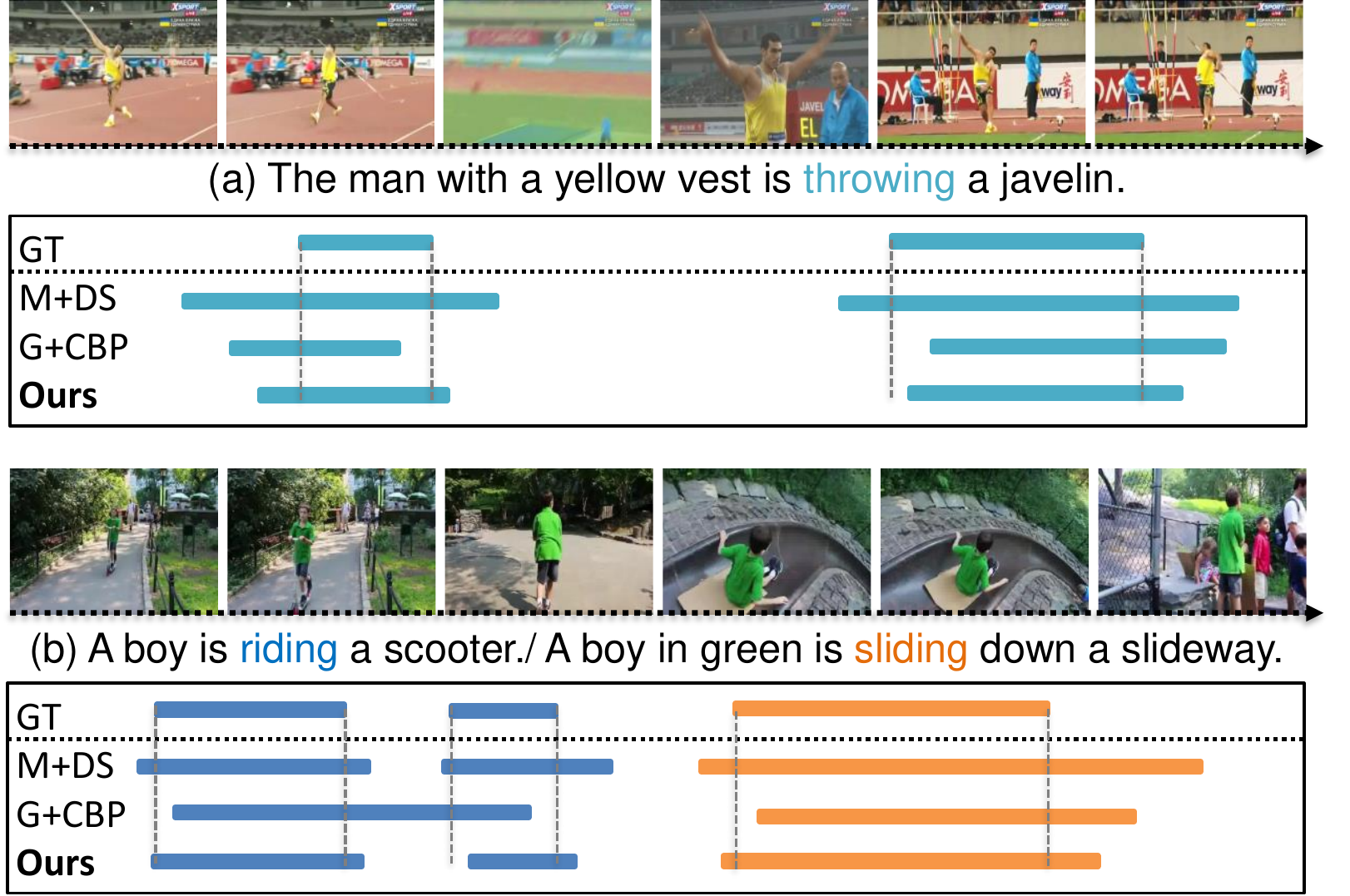}
    \caption{Comparison of temporal localization results between baselines and DSTG showing that our method performs better than the anchor-based method.}
    \label{fig:TG_vis}
    \vspace{-3mm}
\end{figure}

\subsubsection{Generic Visual Grounding}
\begin{table*}[t]
\setlength{\tabcolsep}{6.8pt}
\caption{Comparisons of baseline methods and the proposed DSTG on the val and test sets of GVG benchmark. We fulfill baselines with representive methods for visual grounding for a thorough comparison.} 
\centering
\begin{tabular}{l cccc cccc}
    \toprule
    \multirow{2}{*} {Method}& \multicolumn{4}{c}{\begin{tabular}[c]{@{}c@{}} \textbf{Val set}\end{tabular}} &
    \multicolumn{4}{c}{\begin{tabular}[c]{@{}c@{}} \textbf{Test set}\end{tabular}} \\ 
    \cline{2-9}
    {}& m\_vIoU & vIoU$_{@0.3}$ & vIoU$_{@0.5}$ & vIoU$_{@0.7}$
    & m\_vIoU & vIoU@$_{@0.3}$ & vIoU$_{@0.5}$ & vIoU$_{@0.7}$ \\ 
    \midrule
    Random Anchor             & 8.7 & 16.8 & 11.3 & 7.5 
                                    & 8.2 & 15.5 & 10.7 & 7.1 \\ 
    Mattnet~\cite{yu2018mattnet}+D.S.   & 11.5 & 17.3 & 13.2 & 10.7 
                                    & 11.7 & 17.7 & 13.0 & 10.5 \\
    Matt.+CTRL~\cite{gao2017tall}   & 12.1 & 17.8 & 14.6 & 11.4 
                 & 12.2 & 17.6 & 14.9 & 11.5 \\
    
    Matt.+TGN~\cite{chen2018tgn}  & 16.4 & 23.3 & 18.1 & 14.7 
                               & 16.8 & 23.5 & 18.6 & 14.8 \\
    Matt.+CBP~\cite{wang2018videos}  & 17.7 & 24.2 & 19.3 & 16.3 
                               & 17.4 & 23.9 & 19.2 & 16.3 \\
    A-ATT~\cite{deng2018visual}+CBP     & 21.2 & 27.3 & 21.3 & 18.4 & 20.1 & 27.6 & 22.4 & 18.6 \\
    \midrule
    DGA~\cite{liu2019learning}+CTRL  & 25.6 & 32.2 & 28.0 & 22.8 & 25.2 & 31.9 & 28.0 & 22.3 \\
    DGA~\cite{chen-2019-weakly}+CBP  & 26.3 & 32.6 & 28.6 & 23.1 & 26.5 & 32.7 & 28.8 & 23.4 \\ 
    WSSTG+CBP & 28.7 & 34.0 & 30.3 & 26.2 & 28.3 & 35.4 & 30.4 & 27.5 \\ 
    \rowcolor{mygray1}
    DSTG w/o STCR & 30.2 & 35.2 & 31.8 & 27.6 & 29.9 & 36.3 & 31.7 & 28.1 \\
    \rowcolor{mygray2}
    \textbf{DSTG} & \textbf{32.7} & \textbf{38.2} & \textbf{34.2} & \textbf{30.5} 
       & \textbf{32.9} & \textbf{38.4} & \textbf{34.5} & \textbf{30.7} \\
    \bottomrule
\end{tabular}
\label{tab:gvg_result}
\vspace{-3mm}
\end{table*}

\begin{table}[t]
    \caption{Comparison of results with baselines for visual grounding and temporal localization on GVG dataset.}
    \centering
    \setlength{\tabcolsep}{7.5pt}
    \begin{tabular}{lcccc} 
    \toprule
     & m\_vIoU & vIoU$_{@0.5}$ & m\_tIoU  & tIoU$_{@0.5}$\\
    \midrule
    R.A.     & 12.1 & 14.6 & 15.4 & 13.7\\
    CTRL     & 18.3 & 19.8& 22.0 & 24.5\\
    TGN      & 21.0 & 23.3& 27.6 & 29.2\\
    CBP      & 24.2 & 26.5& 32.8 & 33.6\\
    \textbf{DSTG}   & \textbf{32.7} & \textbf{34.2} & \textbf{38.2} & \textbf{40.3}\\
    \bottomrule
    \end{tabular}
    \vspace{-3mm}
    \label{tab:temporal_performance}
\end{table}

\begin{table}[t]
    \caption{Performance comparison on GVG dataset splits into VG$_{easy}$: single-object within a single segment, SG$_{hard}$: multiple objects, TG$_{hard}$: multiple segments.}
    \centering
    \setlength{\tabcolsep}{6.3pt}
    \begin{tabular}{l cccc} 
    \toprule
    {} & GVG & VG$_{easy}$ & SG$_{hard}$ & TG$_{hard}$\\
    \midrule
    R.A.           & 8.7  & 12.1 & 8.5  & 6.4  \\
    Matt.+CTRL     & 12.1 & 21.1 & 14.2 & 13.4 \\
    Matt.+CBP      & 17.7 & 22.4 & 14.6 & 15.1 \\
    WSSTG+CBP      & 26.3 & 32.3 & 21.5 & 19.7 \\
    \textbf{DSTG}   & \textbf{32.7} & \textbf{35.8} 
                      & \textbf{27.4} & \textbf{23.2} \\
    \bottomrule
    \end{tabular}
    \vspace{-3mm}
    \label{tab:hard_splits}
\end{table}

\noindent\textbf{Baselines.} For a fair and thorough comparison, we extend the existing state-of-the-art methods for generic visual grounding. We extracted I3D features of RoI in videos of GVG with Inception 3D-based architecture which is pretrained on Kinetics~\cite{carreira2017quo}. For temporal localization, we first implement a naive method with Random Anchor by randomly selecting the temporal window anchor in the video. We adopt the anchor-based methods. CTRL~\cite{gao2017tall} is also based on the sliding window anchors to localize the temporal segment. TGN~\cite{chen2018tgn} and CBP~\cite{wang2020temporally} are methods that profit the contextual representation to regress a more accurate temporal boundary. For spatial localization, two types of methods are compared here: modular-based~\cite{yu2018mattnet, deng18AATT}  and  graph-based methods~\cite{yang2019dynamic, chen-2019-weakly}. 

\noindent\textbf{Performance.} We present results in Table~\ref{tab:gvg_result} on both valid set and test set of GVG. It is clear that the grounding model has to retrieve videos not only based on the global view, but also exploit the fine-grained information to accurately localize the required target. 
The module-based methods independently ground sentences in every frame and achieve worse performance than DGA and WSSTG methods, validating the temporal object dynamics across frames are vital for spatio-temporal video grounding.
For spatio-temporal grounding, our DSTG outperforms all baselines on both declarative and interrogative sentences with or without temporal ground truth, which suggests our cross-modal spatio-temporal graph reasoning can effectively capture the object relationships with temporal dynamics and our spatio-temporal localizer can retrieve the object tubes precisely.

\noindent\textbf{Temporal grounding.} Observing results in Table~\ref{tab:temporal_performance}, DSTG achieves a better performance than the frame-level localization method CTRL and CBP, demonstrating the spatio-temporal region modeling is effective to determine the temporal
boundaries of object tubes. It is proved that the anchor-based methods are scaffolded by the inaccurate temporal boundaries.

\begin{figure}[t]
    \centering
    \includegraphics[width=\hsize]{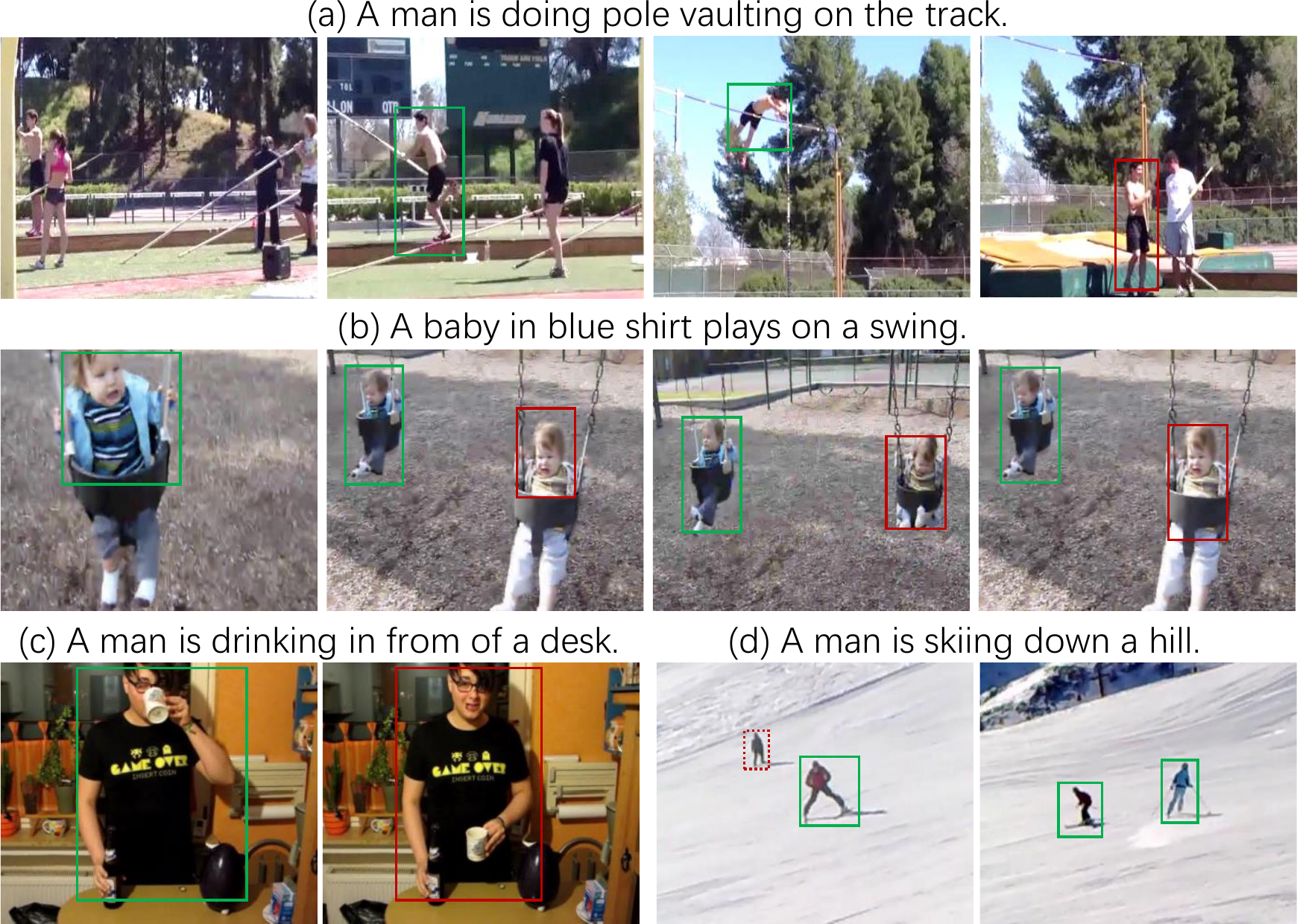}
    \caption{Some failure cases from GVG, marked with red bounding-boxes. Cases in (a) and (c) confuse the model with objects with similar appearance but different motion; (b) shows distractors with similar motion but difference appearance; (d) fails in capturing small-scale target with invisible motion change.}
    \label{fig:single_samples}
    \vspace{-3mm}
\end{figure}

\noindent\textbf{Hard splits.} 
GVG covers diverse grounding cases with the referring expressions and the targets in the untrimmed videos. Beyond the general pair-wise cases that have been well explored by previous practices, we would like to highlight several challenging yet practical cases. 
Results in Table~\ref{tab:hard_splits} are reported on GVG dataset by splitting the valid set and test set into VG$_{easy}$: single-object within a single segment, SG$_{hard}$: multiple objects, TG$_{hard}$: multiple segments, with a ratio of 1:0.83:0.6. Performance on the hard splits drops dramatically especially on TG$_{hard}$.
Some failure cases are presented in Figure~\ref{fig:single_samples}.

\begin{figure*}[t]
    \centering
    \includegraphics[width=\hsize]{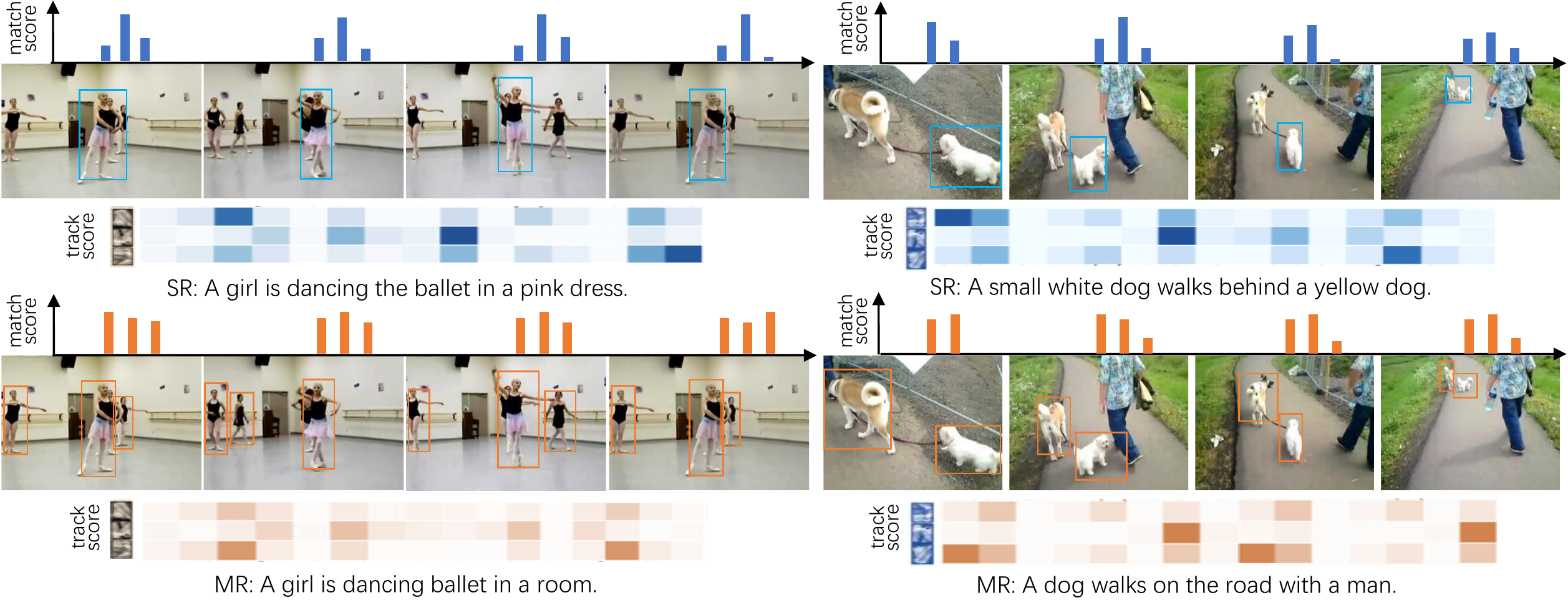}
    \caption{Visual grounding results in the challenging cases where exist multiple targets for grounding. Grounding results are shown together with the matching scores and tracking scores.}
    \label{fig:multi_samples}
    \vspace{-3mm}
\end{figure*}

\subsection{Qualitative Analysis}
\begin{figure}[t]
    \centering
    \includegraphics[width=\hsize]{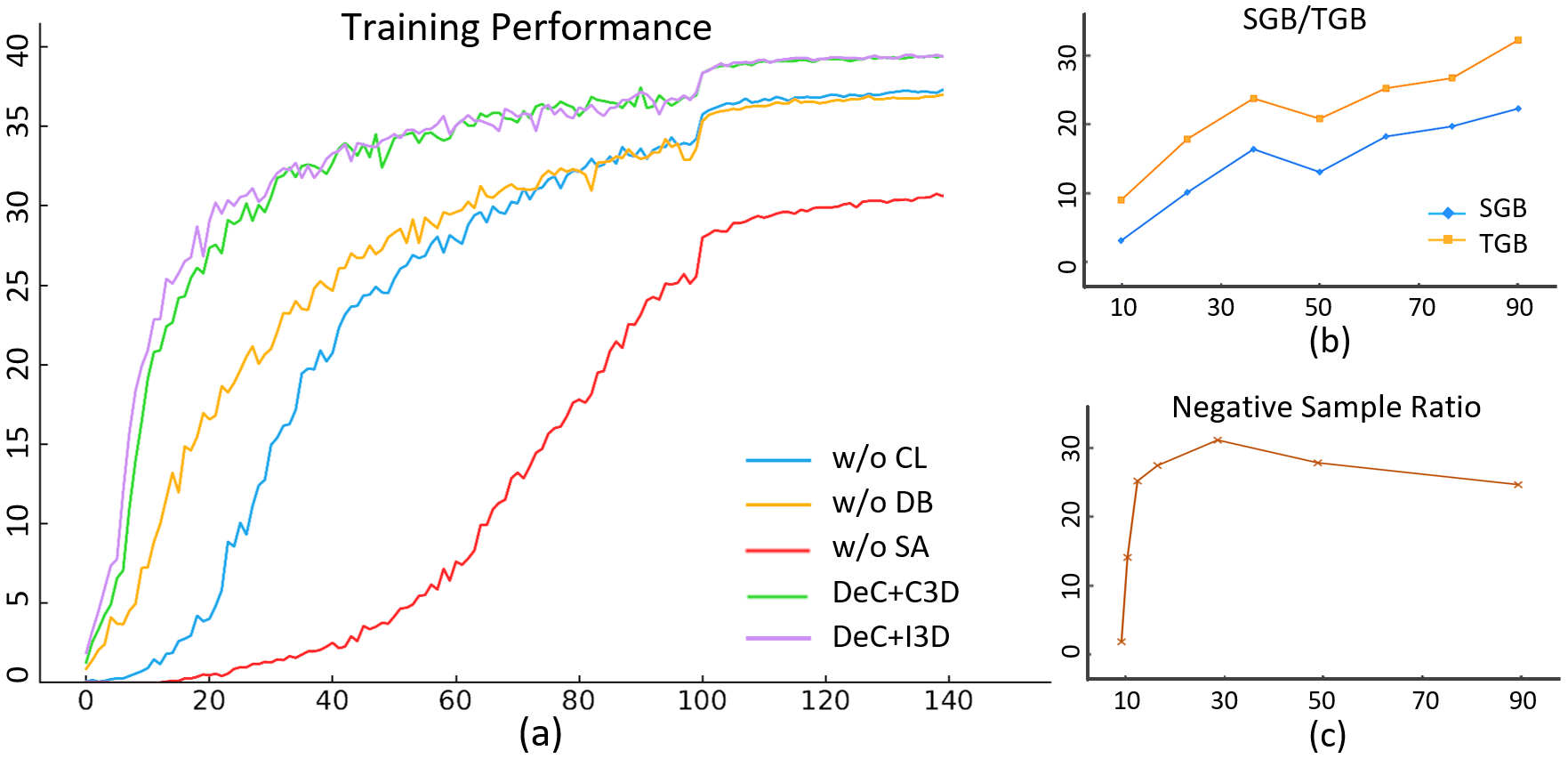}
    \caption{Plot in (a) shows training performance with modules ablated from DSTG. (b) presents evaluation scores on m\_vIoU with only spatial graph branch/temporal graph branch. In (c), we show the effect of negative sample ratio against positive pairs in STCR. }
    \label{fig:plot_ablation}
    \vspace{-3mm}
\end{figure}

\noindent\textbf{Module ablation.}
We show the advantage of each module in the proposed DSTG against the baseline methods in Table~\ref{tab:ablation_module}.
SGB shows visual grounding results with a single spatial branch by taking the referring expression as an integral embedding. It performs below average without learning from the temporal aspects and reaches m\_vIoU of 15.9\%. Besides, the effect of contrastive learning is well-demonstrated in the ablation of SC and TC in the proposed method without sampling the spatial or temporal distractors. SCL improves the score of SPG by 0.5\% and TCL brings an increase of 1.7\% in m\_vIoU compared to SGB+TGB.
Furthermore, the influence of FPS and negative ratio in data sampling are verified and shown in Figure~\ref{fig:plot_ablation}. 
The performance is optimal when FPS equals 6. DSTG achieves the best performance when the ratio of negative samples against the positive ones is set to 5 in the STCR strategy.

\begin{table}[t]
    \centering
    \caption{Results of module ablation in DSTG. SGB/TGB are spatial branch and temporal branch in the encoder. SCL/TCL represent the spatial and temporal consistency in STCR. SA/CA denote self-attenion layers/ cross-attention in decoder.}
    \setlength{\tabcolsep}{7.2pt}
    \begin{tabular}{ccccccc} 
    \toprule
    \multicolumn{2}{c}{\begin{tabular}[c]{@{}c@{}} encoder \end{tabular}} &
    \multicolumn{2}{c}{\begin{tabular}[c]{@{}c@{}} STCR \end{tabular}} &
    \multicolumn{2}{c}{\begin{tabular}[c]{@{}c@{}} attention\end{tabular}} &
    \multirow{2}{*}{m\_vIoU}  \\ \cline{1-6}
    SGB & TGB & SCL & TCL & SA & CA & \\
    \midrule
    $\checkmark$  &    &    &  &    &    & 15.9\\
    $\checkmark$  &  $\checkmark$  &    & &    &    & 16.4\\
    $\checkmark$  &  &  $\checkmark$  &  &    &    & 25.6\\
     &  $\checkmark$  &   & $\checkmark$ &    &      & 27.3\\
    $\checkmark$   &  $\checkmark$  &   $\checkmark$ &  $\checkmark$  & &  & 28.2\\
    $\checkmark$  &  $\checkmark$  & $\checkmark$  &  $\checkmark$   &   $\checkmark$  &   &  30.5\\
    $\checkmark$  & $\checkmark$  & $\checkmark$  &  $\checkmark$  &  $\checkmark$  & $\checkmark$  & 32.7\\
    \bottomrule
    \end{tabular}
    \vspace{-3mm}
    \label{tab:ablation_module}
\end{table}

\noindent \textbf{Appearance cues matters for temporal localization.}
With the spatial and temporal branches, our DSTG is able to focus separately on the spatial characteristics and the temporal patterns. The spatial representation comprises more details about the object attributes, \eg, color, outfit, relationship with the context. Meanwhile, the temporal representation contains the patterns of the motion to assist the temporal understanding for  distinguishing the target from distractors.

\noindent \textbf{Motion cues matters for spatial grounding.}
The spatial-temporal grounding in the untrimmed video extends the spatial grounding into the temporal dimension. Previous approaches deal with the visual grounding task by taking the video content as a whole. Instead, we propose to discover the motion representation separately from the spatial representation.  Some \emph{multiple referring cases} in Figure~\ref{fig:multi_samples} shows that the model counts a lot on the temporal representation to distinguish from the spatial distractors.

\noindent \textbf{Mutual promotion with panoramic cues.} 
With the spatial and temporal branches, our DSTG is able to focus separately on the spatial characteristics and the temporal patterns. The spatial representation comprises more details about the object attributes, \eg, color, outfit, relationship with the context. Meanwhile, the temporal representation contains the patterns of the motion to assist the temporal understanding for distinguishing the target from distractors.



\section{Conclusion}
We interrogate the visual grounding task and reconsider it in a more general form, where referring cases are with large diversity and complexity in space and time. For a better balance between space and time, we resolve it by decomposing the spatial and temporal representations. The critical challenges exist in the spatial and temporal distractors with similar attributes. A proposal-free spatio-temporal contrastive routing strategy is elaborated. We will further explore this new benchmark with more details and reveal the potential development for visual grounding in the future.

{\small
\bibliographystyle{ieee_fullname}
\bibliography{egbib}
}

\end{document}